%% file: arxiv_paper_main.tex
\pdfoutput=1
\documentclass{article}

\usepackage[final,nonatbib]{nips_2018}

\usepackage[T1]{fontenc}    

\usepackage{amsmath, amssymb, amsfonts, bm}
\usepackage{amsthm}
\usepackage{booktabs}       
\usepackage{mathtools}
\usepackage{nicefrac}       
\usepackage[usenames, dvipsnames]{color}
\usepackage{multirow}
\usepackage{hyperref}

\usepackage{floatrow}
\usepackage{wrapfig}
\usepackage{multicol}

\usepackage{array}
\newcolumntype{C}{@{\extracolsep{0.1cm}}c@{\extracolsep{0pt}}}%

\def\dir{}

\def\sec#1{Sec.~\ref{sec:#1}}

\def\fig#1{Fig.~\ref{#1}}
\def\Fig#1{Fig.~\ref{#1}}
\def\tableref#1{Table~\ref{#1}}

\newcommand{\cw}{\mathbf{cw}}
\newcommand{\cc}{\mathbf{c}}
\newcommand{\kb}{\mathbf{k}}
\newcommand{\mem}{\mathbf{m}}
\newcommand{\rr}{\mathbf{r}}

\title{On transfer learning using a MAC model variant}

\author{
	\textbf{\null\hfill Vincent Marois \hfill T.S. Jayram \hfill Vincent Albouy  \hfill\null}\\
	\textbf{\null\hfill Tomasz Kornuta \hfill Younes Bouhadjar \hfill Ahmet S. Ozcan \hfill\null}\\
	IBM Research AI, Almaden Research Center, San Jose, USA\\
	\texttt{\{vmarois,jayram,tkornut,byounes,asozcan\}@us.ibm.com}\\
	\texttt{\{vincent.albouy\}@ibm.com}\\
}

\begin{document}

\maketitle

\begin{abstract}
We introduce a variant of the MAC model (Hudson and Manning, ICLR 2018) with a simplified set of equations that achieves comparable accuracy, while training faster. We evaluate both models on CLEVR and CoGenT, and show that, transfer learning with fine-tuning results in a 15 point increase in accuracy, matching the state of the art. Finally, in contrast, we demonstrate that improper fine-tuning can actually reduce a model's accuracy as well.
\end{abstract}
\input{introduction}
\input{datasets}

\input{mac}
\input{smac}

\input{experiments}
\input{comparison}
\input{conclusion}

\bibliographystyle{abbrv}
\bibliography{\dir/vigil_bibliography}

\newpage
\appendix
\input{appendix}

\end{document}

%% file: introduction.tex
\section{Introduction}
Reasoning over visual inputs is a fundamental characteristic of human intelligence.
Reproducing this ability with artificial systems is a challenge that requires learning relations and compositionality~\cite{hu2017learning, johnson2017inferring}. The Visual Question Answering (VQA)~\cite{antol2015vqa,malinowski2014towards,wu2017visual} task has been tailored to benchmark this type of reasoning, combining natural language processing and visual recognition.

Many approaches have been explored, such as modular networks, that combine modules coming from a predefined collection \cite{andreas2016learning,johnson2017inferring, mascharka2018transparency}. Attention mechanisms (\cite{bahdanau2014neural}, \cite{xu2015show}) are also used to guide the focus of the system over the image and the question words.

Several VQA datasets have been proposed (e.g. DAQUAR~\cite{malinowski2014multi}, VQA~\cite{antol2015vqa}); nonetheless, these datasets contain several biases (e.g. unbalanced questions or answers) that are often exploited by systems during learning~\cite{goyal2017making}.
The CLEVR dataset~\cite{johnson2017clevr} was designed to address these issues. The synthetic nature of its images \& questions enables detailed analysis of visual reasoning, and allows for variations, to test a particular ability such as generalization or transfer learning. One variation of CLEVR, called CoGenT (Compositional Generalization Test), measures whether models learn separate representations for color and shape instead of memorizing all possible combinations.

One of the most recent exciting models aiming at solving VQA is called Memory, Attention, and Composition (MAC)~\cite{hudson2018compositional}, which performs a sequence of attention-based reasoning operations. 
though the performance of MAC has been proven, several questions arise:
Does the model really learn relations between objects? 
How does the model represent these relations in its reasoning steps? 
Is the model representing concepts like objects attributes (shape, color, size)?

In this work, we further investigate the interpretability and generalization capabilities of the MAC model.
We propose a new set of equations that simplifies the core of the model (S-MAC). It trains faster and achieves comparable accuracy on CLEVR. 
Second, we show that both models achieve comparable performance with zero-shot learning, when trained on CoGenT-A and tested on CoGenT-B. With fine-tuning however, we obtained a significant 15 points increase in the accuracy, matching state-of-the-art results~\cite{perez2017film, mascharka2018transparency}.
Last, we illustrate using CoGenT-B that, without adequate care, fine-tuning can actually reduce a model's accuracy.

The paper is organized as follows.
In \sec{datasets} we delve into details of both CLEVR and CoGenT datasets, emphasizing their differences.
Next, in \sec{models} we present both MAC and S-MAC models and explain what motivated the formulation of equations for the latter.
\sec{experiments} discusses the most important results of using transfer learning 
on CLEVR and on both versions of CoGenT.
\sec{transfer_learning} focuses on a cherry-picked subset of experiments to highlight our findings; 
see~Appendix~\ref{sec:full_comparison} for the complete set of results comparing MAC and S-MAC models when training/fine-tuning/testing on various datasets.
In \sec{failures} we show examples supporting hypothesis on the lack of disentanglement of concepts in both models.
\sec{comparison} presents results obtained by other state-of-the-art models, and discusses the challenges faced  
with their reproducibility. 
\sec{conclusion} concludes the paper.

%% file: datasets.tex
\section{Description of datasets}
\label{sec:datasets}

Most of the VQA datasets have strong biases. This allow models to learn strategies without reasoning about the visual input~\cite{Santoro2017ASN}.
The CLEVR dataset~\cite{johnson2017clevr} was developed to address those issues and come back to the core challenge of visual QA which is testing reasoning abilities.
CLEVR contains images of 3D-rendered objects; each image comes with a number of highly compositional questions that fall into different categories.
Those categories fall into 5 classes of tasks: Exist, Count, Compare Integer, Query Attribute and Compare Attribute. 
The CLEVR dataset consists of:
\begin{itemize}
\item 	A training set of 70k images and 700k questions,
\item	A validation set of 15k images and 150k questions,
\item	A test  set of 15k images and 150k questions about objects,
\item	Answers, scene graphs and functional programs for all train and val images and questions.
\end{itemize}
Each object present in the scene, aside of position, is characterized by a set of four attributes:
\begin{itemize}
\item 2 sizes: large, small,
\item 3 shapes: square, cylinder, sphere,
\item 2 material types: rubber, metal,
\item 8 color types: gray, blue, brown, yellow, red, green, purple, cyan,
\end{itemize}
resulting in 96 unique combinations.

Along with CLEVR, the authors~\cite{johnson2017clevr} introduced  CLEVR-CoGenT (Compositional Generalization Test, CoGenT in short), with a goal of evaluating how well the models can generalize, learn relations and compositional concepts.
This dataset is generated in the same way as CLEVR with two additional conditions.
As shown in \tableref{tab:cogent_conditions}, in Condition A all cubes are gray, blue, brown, or yellow, whereas all cylinders are red, green, purple, or cyan; in Condition B cubes and cylinders swap color palettes.
For both conditions spheres can be any colors.

\begin{table}[b!]
	\centering
	\begin{tabular}{cccc}
		\toprule
		Dataset        & Cubes              & Cylinders &  Spheres         \\
		\midrule
		CLEVR   &  any color &  any color        &    any color    \\
		CLEVR CoGenT A & gray / blue / brown / yellow  & red / green / purple / cyan       &    any color  \\
		CLEVR CoGenT B  & red / green / purple / cyan &   gray / blue / brown / yellow       &      any color  \\
		\bottomrule
	\end{tabular}
	\caption{Colors/shapes combinations present in CLEVR, CoGenT-A and CoGenT-B datasets.}
	\label{tab:cogent_conditions}
\end{table}

The CoGenT dataset contains:
\begin{itemize}
\item	Training set of 70,000 images and 699,960 questions in Condition A,
\item	Validation set of 15,000 images and 149,991 questions in Condition A,
\item	Test set of 15,000 images and 149,980 questions in Condition A (without answers),
\item	Validation set of 15,000 images and 150,000 questions in Condition B,
\item	Test set of 15,000 images and 149,992 questions in Condition B (without answers),
\item	Scene graphs and functional programs for all training/validation images/questions.
\end{itemize}

%% file: mac.tex
\section{MAC network and our proposed simplification}
\label{sec:models}

The MAC network~\cite{hudson2018compositional} is a recurrent model that performs sequential reasoning, where each step involves analyzing a part of the question followed by shifting the attention over the image.
The core of the model is the MAC cell, supported with an input unit that processes the question and image pair, and output unit which produces the answer.
The input unit  uses an LSTM~\cite{hochreiter1997long} to process the question in a word-by-word manner producing a sequence of \emph{contextual words} and a final \emph{question representation}.
Besides, the input unit utilizes a pre-trained ResNet~\cite{he2016resnet} followed by two CNN layers to extract a feature map (referred to as \emph{knowledge base}) from the image.

\begin{wrapfigure}[10]{r}[5pt]{0.4\textwidth}
	\vspace{-15pt}
	\centering
	\includegraphics[width=\textwidth]{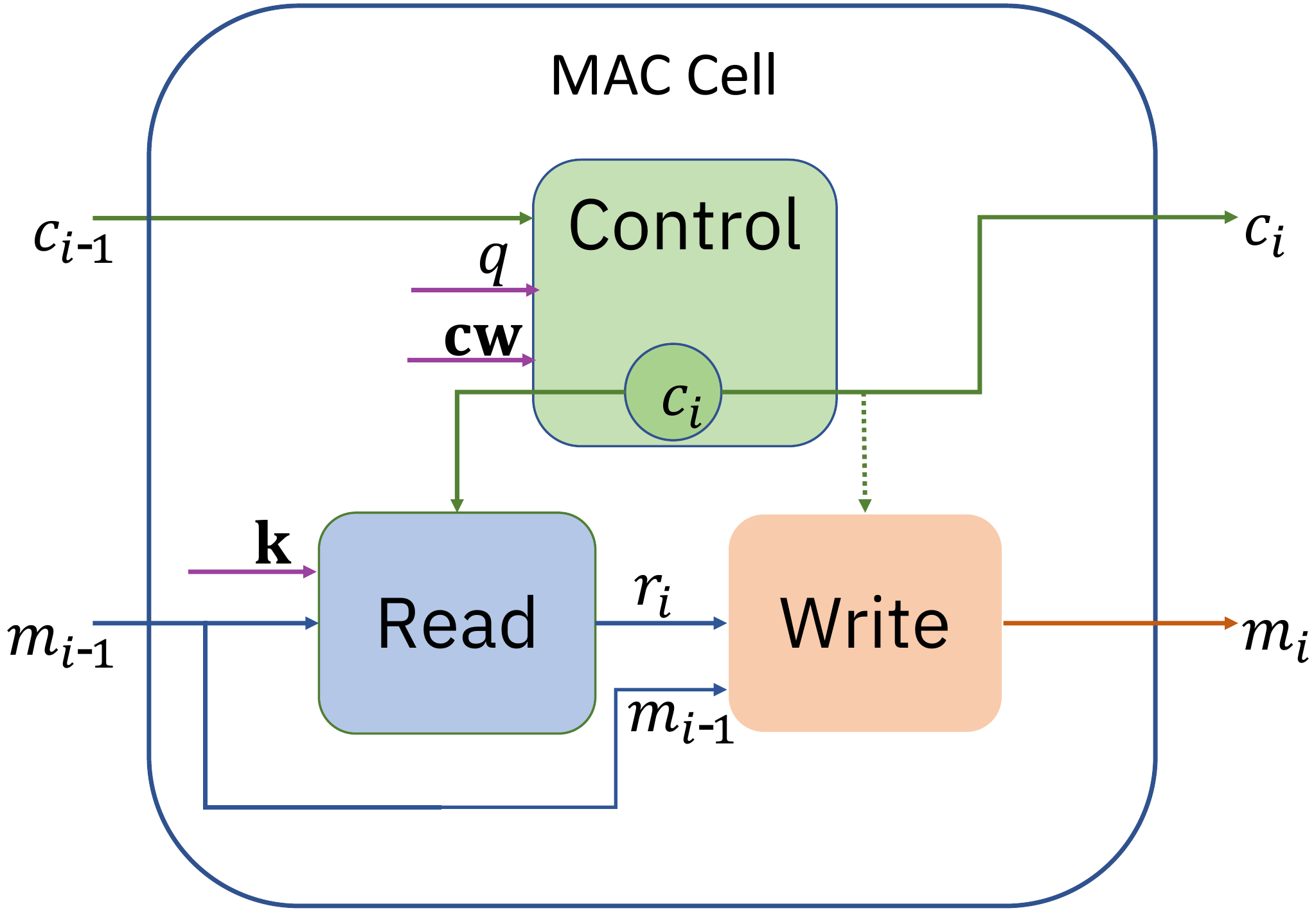}
	\caption{The MAC cell, reproduced on the basis of~\cite{hudson2018compositional}.}
	\label{fig:mac_cell}
	\vspace{-5pt}
\end{wrapfigure}

The MAC cell consists of a control unit, a read unit and a write unit (\Fig{fig:mac_cell}).
The control unit is updating the control state $c_i$ and drives the attention over the list of \emph{contextual words} $\cw$ taking into account the \emph{question representation} $q$.
Guided by $c_i$,  the read unit extracts information from the \emph{knowledge base} $\kb$ and combines it with the previous memory state $m_{i-1}$  to produce the \emph{read vector} $r_i$.
Finally, the write unit integrates $r_i$ and $m_{i-1}$ to update the memory state. Detailed equations are described in the next section.

%% file: smac.tex
\subsection{Simplified MAC network}
Our proposed modification to the MAC network is based on two heuristic
simplifications of the MAC cell. 
First, we observe that, taking the MAC cell equations as a whole, consecutive linear layers (with no activation in-between) can be combined as one linear layer.
Next, we assume that dimension-preserving linear layers are invertible
so as to avoid information loss. 
Applying this principle to the equations, with a careful reorganization,
we can apply a single linear layer to the knowledge base \emph{prior}
to all the reasoning steps and work with this \emph{projected} knowledge base
as-is throughout the reasoning steps.


\noindent\textit{Notation.}
For each knowledge base $\kb$ of dimension $H \times W \times d$, let $\kb_{hw}$ be the $d$-dimensional vector indexed by 
$h \in \{1,2,\dots,H\}$ and $w \in \{1,2,\dots, W\}$.
Let $i$ denote the index of the reasoning steps. 

In the description below, the original MAC cell equations are shown on the \emph{left}
while our simplified equations are shown (in color) on the {\color{Plum} \emph{right}}.
The equation numbering is the same as in~\cite{hudson2018compositional}.

\noindent\textbf{Control unit:} 
For both models, in the control unit, the question $q$ is first transformed in each step of 
the reasoning using a \emph{position-aware}
linear layer depending on $i$: $q_i = U_i^{[d \times 2d]} q + b_i^{[d]}$.

\begin{multicols}{2}
	\noindent
	\begin{align*}
	&cq_i = W_{cq}^{[d \times 2d]} [c_{i-1}, q_i] + b_{cq}^{[d]}  \tag{c1} \\
	&ca_{is} = W_{ca}^{[1 \times d]} (cq_i \odot \cw_s) + b_{ca}^{[1]}
	\tag{c2.1}\\
	&cv_{is} = \textrm{softmax}(ca_{is}) \tag{c2.2}\\
	&\cc_i = \sum_s cv_{is} \, \cw_s  \tag{c2.3}
	\end{align*}
	\columnbreak
	{\color{Plum}
	\begin{align*}
	&cq_i = W_{cq}^{[d \times d]} c_{i-1} + q_i  \tag{c1} \\
	&ca_{is} = W_{ca}^{[1 \times d]} (cq_i \odot \cw_s)  \tag{c2.1}\\
	&cv_{is} = \textrm{softmax}(ca_{is}) \tag{c2.2}\\
	&\cc_i = \sum_s cv_{is} \, \cw_s  \tag{c2.3}
    \end{align*}}
\end{multicols}

\vskip -0.6cm
\noindent\textbf{Read and write units:}
\begin{multicols}{2}
	\noindent
	\begin{align*}
	&I_{ihw} = (W_{m}^{[d \times d]} \mem_{i-1} + b_{m}^{[d]}) \\
	           & \qquad \quad \odot (W_{k}^{[d \times d]} \kb_{hw} + b_{k}^{[d]}) \tag{r1} \\
	&I'_{ihw} =  W_{I'}^{[d \times 2d]} [I_{ihw},\kb_{hw}]  + b_{I'}^{[d]}  \tag{r2} \\
	&ra_{ihw} = W_{ra}^{[1 \times d]} (\cc_i \odot I'_{ihw}) + b_{ra}^{[1]} \tag{r3.1}\\
	&rv_{ihw} = \textrm{softmax}(ra_{ihw}) \tag{r3.2}\\
	&\rr_i = \sum_s rv_{ihw} \, \kb_{hw}  \tag{r3.3}\\
	&\mem_i = W_{rm}^{[d \times 2d]} [\rr_i, \mem_{i-1}]  + b_{rm}^{[d]} \tag{w1}	
	\end{align*}
	\columnbreak
	{\color{Plum}
	\begin{align*}
	&I_{ihw} = \mem_{i-1} \odot \kb_{hw} \tag{r1} \\ \\
	&I'_{ihw} = W_{I'}^{[d \times d]} I_{ihw} + b_{I'}^{[d]} + \kb_{hw} \tag{r2} \\
	&ra_{ihw} = W_{ra}^{[1 \times d]} (\cc_i \odot I'_{ihw})  \tag{r3.1}\\
	&rv_{ihw} = \textrm{softmax}(ra_{ihw}) \tag{r3.2}\\
	&\rr_i = \sum_s rv_{ihw} \, \kb_{hw}  \tag{r3.3}\\
	&\mem_i = W_{rm}^{[d \times d]} \rr_i + b_{rm}^{[d]} \tag{w1}
	\end{align*}}
\end{multicols}

As seen above, being noticeably simpler, the S-MAC obtains significant reduction in the number of position-independent parameters across all units (see~\tableref{tab:parameters}).
Our experiments demonstrate that this gives us noticeable savings in the training time. However, since the computation time is also dominated by the position-aware layers in the control unit, as well as the input unit, the speedup is not as large as we desire.

\begin{table}[!t]
	\centering
	\begin{tabular}{ccccCcCc}
		\toprule
		Model        & Read Unit               & Write Unit &  Control Unit         \\
		\midrule
		MAC   &  787,969 &  524,800        &    525,313    \\
		simplified MAC & 263,168  & 262,656       &    263,168 \\
		\midrule
		Reduction by [\%]  & 67\%  &   50\%       &      50\%  \\
		\bottomrule
	\end{tabular}
	\caption{Comparison of the number of position-independent parameters in MAC \& S-MAC cells.}
	\label{tab:parameters}
\end{table}

%% file: experiments.tex
\section{Experiments}
\label{sec:experiments}

Our experiments were intended to study MAC's and S-MAC's generalization as well as transfer learning abilities in different settings. We used CLEVR and CoGenT to address these different aspects.
The first experiment studied the training time and the capability of the models to generalize on the same type of dataset that it was trained on. This was used mainly as a baseline for the further experiments that were intended to study how well the transfer learning performed in comparison to the baseline results.
The second experiment studied the capability of the models to succeed in doing transfer learning from domain A to domain B when trained on different combinations of the respective domains. The third experiment was intended to see whether the performance improves if the model could be further trained on a small subset of the dataset from domain B.
\tableref{tab:results} presents the most important results, focusing on S-MAC in particular; see Appendix~\ref{sec:full_comparison} for the entire set.

For all experiments, the initial training procedure is as follows: we train the given model  for 20 epochs on 90\% of the training sets of CLEVR \& CoGenT separately. We keep the remaining 10\% for validating the model at every epoch, and use the original validation sets as test sets.

Our implementation of both MAC models used PyTorch (v0.4.0)~\cite{paszke2017automatic}. We relied on the MI-Prometheus~\cite{kornuta2018accelerating} framework that enables fast experimentation of the cross product of models and datasets\footnote{To reproduce the presented research please follow:  \url{https://github.com/IBM/mi-prometheus/}}.
We used NVIDIA's GeForce GTX TITAN X GPUs. We followed the implementation details indicated in the supplemental material (Sec. A) of the original paper~\cite{hudson2018compositional}, to ensure a faithful comparison.

\subsection{Transfer Learning}
\label{sec:transfer_learning}

\begin{table}[t]
	\caption{CLEVR \& CoGenT accuracies for the MAC \& S-MAC models. The [Training] column indicates wall times and final accuracies on the training set. For fine-tuning, we use 30k samples of the test set, and kept the remainder for testing. The [Fine-tuning] column reports the used sub-set (30k samples) and the final accuracies on this sub-set during training. The [Test] column reports the used set and the obtained test accuracies. If no fine-tuning was done, the whole indicated set was used for testing.}
	\centering
	\begin{tabular}{cccccCcCC}
		\toprule
		\multirow{2}{*}{Model} & \multicolumn{3}{c}{Training} &  \multicolumn{2}{c}{Fine-tuning} & \multicolumn{2}{c}{Test} & \multirow{2}{*}{Row} \\
		\cmidrule{2-4} \cmidrule{5-6} \cmidrule{7-8} 
		& Dataset                & Time [h:m] & Acc [\%]          & Dataset & Acc [\%]  & Dataset & Acc [\%] & \\
		\midrule
		MAC & CLEVR  & 30:52  & 96.70 & --   & --  & CLEVR    & 96.17         & (a) \\
		\cmidrule{1-8}
		\cmidrule{1-8}
				
		 \multirow{13}{*}{S-MAC} & CLEVR  & 28:30  & 95.82 & --   & --  & CLEVR    & 95.29         & (b)  \\
		 \cmidrule{2-4} \cmidrule{5-6} \cmidrule{7-8} 
		
		& CoGenT-A  & 28:33   & 96.09 &  --  &  --  & CoGenT-A & 95.91        & (c)  \\
		\cmidrule{2-4} \cmidrule{5-6} \cmidrule{7-8}

		& \multirow{2}{*}{CLEVR}  & \multirow{2}{*}{28:30}  & \multirow{2}{*}{95.82} & \multirow{2}{*}{--}   & \multirow{2}{*}{--}  &   CoGenT-A    &  95.47  & (d) \\
		\cmidrule{7-8} 
		&                        &   &              &     &                               & CoGenT-B   &  95.58  & (e)\\		
				
		\cmidrule{2-4} \cmidrule{5-6} \cmidrule{7-8} 
		& \multirow{4}{*}{CoGenT-A}   & \multirow{4}{*}{28:33}   & \multirow{4}{*}{96.09}  &  \multirow{1}{*}{--}  &  \multirow{1}{*}{--}   & CogenT-B & 78.71        & (f)  \\
		\cmidrule{5-6} \cmidrule{7-8} 
		&                             &                                         &    &   \multirow{2}{*}{CoGenT-B}         &       \multirow{2}{*}{96.85}          & CoGenT-A &  91.24        & (g) \\
		\cmidrule{7-8} 
		&                             &                                         &       &         &                & CoGenT-B &    94.55     & (h)  \\

		\cmidrule{2-4} \cmidrule{5-6} \cmidrule{7-8} 
& \multirow{2}{*}{CLEVR}  & \multirow{2}{*}{28:30}  & \multirow{2}{*}{95.82} &   \multirow{2}{*}{CoGenT-B}         &       \multirow{2}{*}{97.67}          & CoGenT-A &  92.11       & (i) \\
\cmidrule{7-8} 
&                             &                                         &       &         &                & CoGenT-B &    92.95    & (j)  \\

		\bottomrule
	\end{tabular}
	\label{tab:results}
\end{table}

For the CLEVR dataset, the training wall time of MAC (row a) is consistent with what is reported in the original paper (roughly 30h of training for 20 epochs).
S-MAC trains faster, showing a decrease of 10.5\% in wall time (row b), due to the reductions in the 
number of parameters as shown earlier.
This was consistently observed across the other experiments as well.

Turning to the generalization performance (row a),  MAC on CLEVR yields an accuracy of 96.17\%, which is 
taken as a reference experiment. S-MAC reaches an accuracy of 95.29\% on CLEVR (row b), indicating that 
the simplifications did not hinder its generalization capability.
Similar performance was observed for generalization on CoGenT-A (row c).

Before fine-tuning, we wanted to estimate the best upper bounds on accuracy that we could possibly get by doing transfer learning.
As both CoGenT datasets contain complementary subsets of colors/shapes  combinations present in CLEVR, 
we evaluated CLEVR-trained models on the CoGenT datasets.
Even though the CoGenT datasets were generated using more restricted parameters, the models obtained nearly equal accuracy (rows d-e).

Evaluating S-MAC on CoGenT shows that, similar to~\cite{johnson2017inferring, mascharka2018transparency}, the score is worse on CoGenT-B (zero-shot learning, row f) than CoGenT-A after training on CoGenT-A data only (generalization, row c).

Following \cite{johnson2017inferring, perez2017film}), we then fine-tune S-MAC using 3k images and 30k questions from the CoGenT-B data (for 10 epochs), and re-evaluate it on both conditions. This enables much higher accuracy on CoGenT-B, of at least a 15 points increase (row h). Performance on CoGenT-A is slightly worse, dropping by 4 points (row g). This seems to indicate that S-MAC is able to learn new combinations of shape \& color without forgetting the ones it learned during the initial training. 

To study the pitfalls of fine-tuning, we conducted a final set of experiments, where we fine-tuned a CLEVR-trained S-MAC model on CoGenT-B for 10 epochs, as before. Surprisingly, this operation handicapped the generalization of the model not only on CoGenT-A (row i), but also on CoGenT-B (row j, a 3 point drop compared to row e). This highlights the delicate nature of fine-tuning, with respect to the correlation between the datasets. This warrants further investigations.

\subsection{Illustration of failures of MAC on CLEVR}
\label{sec:failures}

Following the evaluation of MAC on CoGenT-B, we built a tool which helped us visualizing the attention of the model over the question words and the image, and thus provide insight on some cases of failure.

\begin{figure}[htbp]
	\centering
	\includegraphics[width=\textwidth]{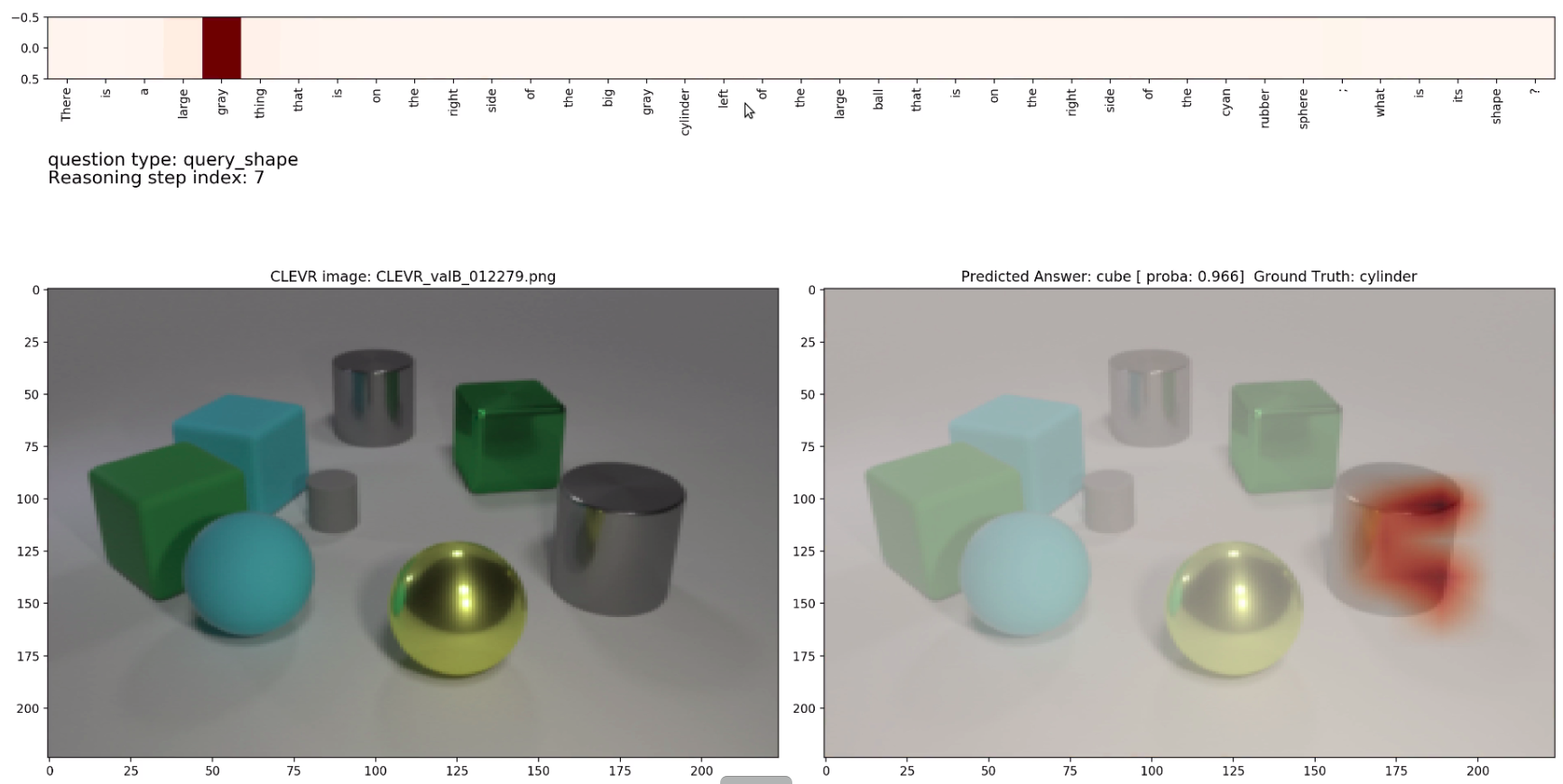}
	\caption{The question reads as: \textit{There is a large gray thing that is on the right side of the big gray cylinder left of the large ball that is on the right side if the cyan rubber sphere; what is its shape?}}
	\label{fig:fail_mac_shape}
\end{figure}

\fig{fig:fail_mac_shape} presents a question where the model is asked about the shape of the leftmost gray cylinder. The model correctly finds it, as we can see from its visual attention map, and appears to refer to it using its color (\textit{gray}), as we can see from the attention of the question words. Yet, it defaults to predicting the shape as \textit{cube}, because it never saw gray cylinders during training, but instead saw gray cubes.

\begin{figure}[htbp]
	\centering
	\includegraphics[width=\textwidth]{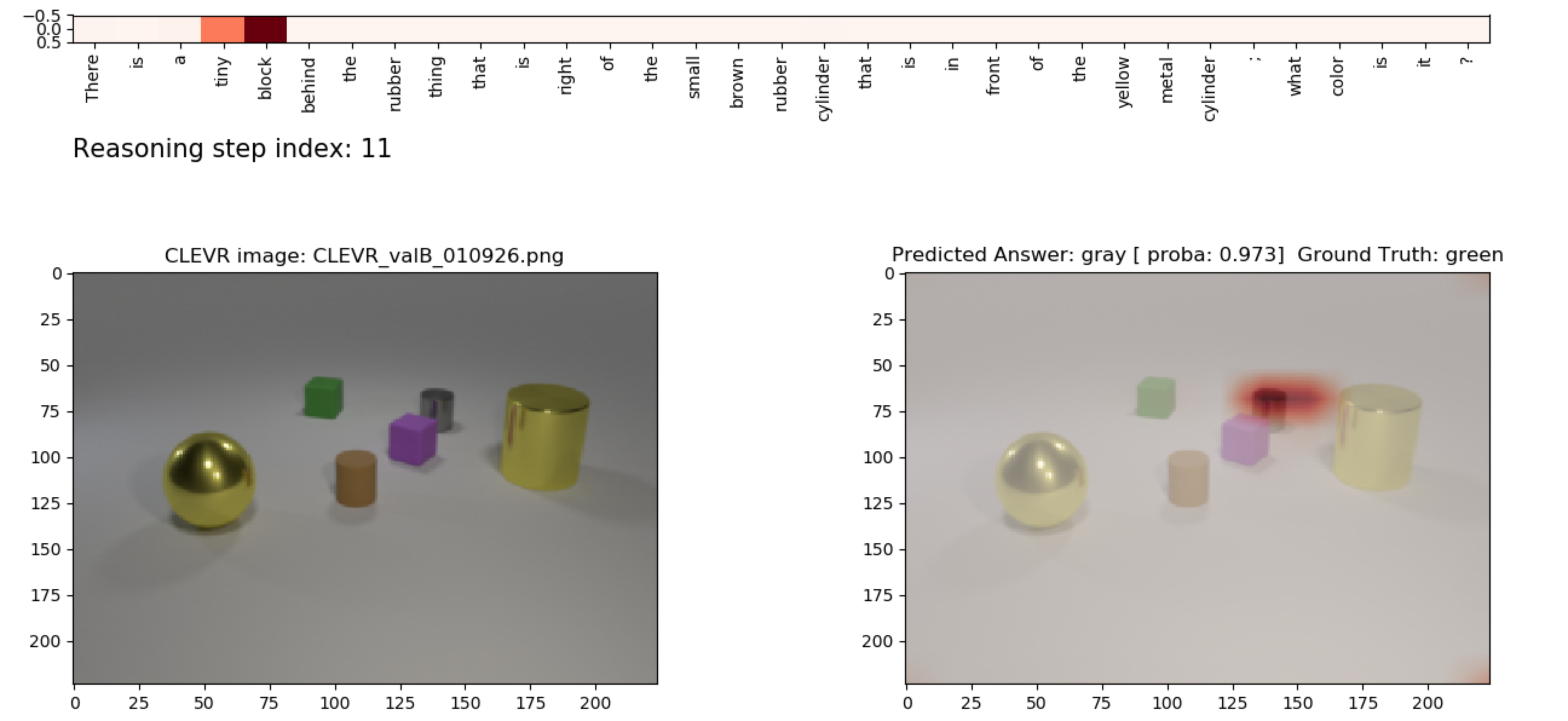}
	\caption{The question reads as: \textit{There is a tiny block behind the rubber thing that is right if the small brown rubber cylinder that is in front of the yellow metal cylinder; what color is it?}}
	\label{fig:fail_mac_color}
\end{figure}

\fig{fig:fail_mac_color} presents a similar case, where the model is questioned about the color of the green cube at the back. MAC misses that object, and instead focuses on the nearby gray cylinder. We can hypothesize that MAC missed the green cube as it did not see this combination during training, and thus defaulted to a combination that it knows.

Those examples indicate that MAC did not correctly separate the concept of shape from the concept of color, but have a better understanding of the colors (as it found the object of interest in \fig{fig:fail_mac_shape} by its color). This could come from that fact that the shape \textit{sphere} is associated with all possible colors in the dataset. 

%% file: comparison.tex
\subsection{Comparison of generalization capabilities with other models}
\label{sec:comparison}

In this section, we present a comparison of our results on generalization capabilities with selected state-of-the-art models.
In particular, we focused on three papers reporting state-of-the-art accuracies on CoGenT. Those papers introduce the following models: PG+EE~\cite{johnson2017inferring}, FiLM~\cite{perez2017film} and TbD~\cite{mascharka2018transparency}.
Deeper analysis of those papers revealed that it is likely that different authors used different sets for reporting the scores, which questions the correctness of the comparison.
We find that the problem results from the fact that ground truth answers for the test sets are not provided along with these sets; thus subsets of the validation sets were sometimes used for testing. 
The results of our research are presented in \tableref{tab:generalization_comparison}, where we shortened the names of the datasets.
For instance, \textbf{A Test Full} means the use of the whole \textbf{CoGenT Condition A Test set}, whereas \textbf{B Valid 30k} indicates the use of 30.000 samples from \textbf{CoGenT Condition B Validation set}.
Question marks indicate that the paper does not provide enough information; thus the indicated sets are the ones we assumed were used.

\begin{table}[!h]
	\centering
	\begin{tabular}{cCcCcCc}
		\toprule
		Model & \multicolumn{2}{c}{Training} &    \multicolumn{2}{c}{Fine-tuning} &   \multicolumn{2}{c}{Test} \\		
		\cmidrule{2-3} \cmidrule{4-5}\cmidrule{6-7}
		(source)& CoGenT set & Acc [\%]  & CoGenT set & Acc [\%]  & CoGenT set~ & Acc [\%] \\
		
		\midrule				
		& \multirow{5}{*}{A Train Full?}   & \multirow{5}{*}{N/A}  & \multirow{2}{*}{--} & \multirow{2}{*}{--}  &   A Test Full    &   96.6  \\
		\cmidrule{6-7} 
		PG+EE &   &    &   &    & B Test Full?    &   73.7  \\
		\cmidrule{4-5}\cmidrule{6-7}
		(\cite{johnson2017inferring}) &  &    & \multirow{2}{*}{B Train 30k?}  & \multirow{2}{*}{N/A}     & A Test Full    &   76.1 \\
		\cmidrule{6-7} 
		&   &    &   &    & B Test Full    &   92.7  \\
		
		\midrule				
		CNN+GRU+FiLM & \multirow{5}{*}{A Train Full?}   & \multirow{5}{*}{N/A}  & \multirow{2}{*}{--} & \multirow{2}{*}{--}  &   A Valid Full?    &  98.3   \\
		\cmidrule{6-7} 
		0-Shot &   &    &   &    & B Valid 120k    &   78.8  \\
		\cmidrule{4-5}\cmidrule{6-7}
		(\cite{perez2017film}) &  &    & \multirow{2}{*}{B Valid 30k}  & \multirow{2}{*}{N/A}     & A Valid Full?    & 81.1  \\
		\cmidrule{6-7} 
		&   &    &   &    & B Valid 120k    &  96.9  \\

		\midrule				
		& \multirow{5}{*}{A Train Full?}   & \multirow{5}{*}{N/A}  & \multirow{2}{*}{--} & \multirow{2}{*}{--}  &   A ?    &  98.8   \\
		\cmidrule{6-7} 
		TbD + reg &   &    &   &    & B ?    &  75.4   \\
		\cmidrule{4-5}\cmidrule{6-7}
		(\cite{mascharka2018transparency}) &  &    & \multirow{2}{*}{B Valid 30k}  & \multirow{2}{*}{N/A}     & A ?    &  96.9 \\
		\cmidrule{6-7} 
		&   &    &   &    & B ?   &  96.3  \\

		\midrule				
		& \multirow{5}{*}{A Train 630k}   & \multirow{5}{*}{97.02}  & \multirow{2}{*}{--} & \multirow{2}{*}{--}  &   A Valid Full    &     96.88 \\
		\cmidrule{6-7} 
		MAC &   &    &   &    & B Valid Full   &  79.54   \\
		\cmidrule{4-5}\cmidrule{6-7}
		(our results) &  &    & \multirow{2}{*}{B Valid 30k}  & \multirow{2}{*}{97.91}     & A Valid Full    &  92.06 \\
		\cmidrule{6-7} 
		&   &    &   &    & B Valid 120k    &   95.62 \\
		
		\midrule				
		& \multirow{5}{*}{A Train 630k}   & \multirow{5}{*}{96.09}  & \multirow{2}{*}{--} & \multirow{2}{*}{--}  &   A Valid Full    &     95.91 \\
		\cmidrule{6-7} 
		S-MAC &   &    &   &    & B Valid Full   &  78.71   \\
		\cmidrule{4-5}\cmidrule{6-7}
		(our results) &  &    & \multirow{2}{*}{B Valid 30k}  & \multirow{2}{*}{96.85}     & A Valid Full    &  91.24 \\
		\cmidrule{6-7} 
		&   &    &   &    & B Valid 120k    &   94.55 \\
		
		\bottomrule
	\end{tabular}
	\caption{Generalization capabilities of selected state-of-the-art models.}
	\label{tab:generalization_comparison}
\end{table}

\subsubsection{The PG+EE model and training methodology}
The PG+EE (Program Generator and Execution Engine)~\cite{johnson2017inferring} model is composed of two main modules:
a Program Generator constructing an explicit, graph-like representation of the reasoning process, and an Execution Engine executing that program and producing an answer. 
Both modules are implemented by neural networks, and were trained using a combination of backpropagation and REINFORCE~\cite{williams1992simple}.

The authors inform that in the first step they trained their models on Condition A, and tested them on both conditions. 
Next, they fine-tuned these models on Condition B using 3K images and 30K questions, and again tested on both conditions.
We could not ascertain which sets they used for fine-tuning (as Condition B lacks a training set).
Being the authors of the CLEVR and CoGenT datasets, it is possible that they generated specific sets. Additionally, as they own the ground truth answers for the test sets, we take as granted that they reported the accuracies on both Condition A and Condition B test sets, despite they explicitly didn't say that in the paper.

\subsubsection{The FiLM model and training methodology}

Feature-wise Linear Modulation (in short, FiLM)~\cite{perez2017film} is an optional enhancement of a neural network model.
The idea is to influence the behavior of existing layer(s) by introducing feature-wise affine transformations which are conditioned on the input.
A model composed of CNNs and a GRU with FiLM-enhanced layers achieved state-of-the-art results on both CLEVR and CoGenT, showing improvement over the PG+EE model.

Nonetheless, the authors indicate in the paper that the accuracy reported for Condition B after fine-tuning was obtained on the CoGenT Condition B Validation set, excluding the 30k samples which were used for fine-tuning.
This suggests that they probably reported scores for Condition A on the validation, not test set as well.

\subsubsection{The TbD model and training methodology}

The TbD (Transparency by Design) network was introduced in~\cite{mascharka2018transparency}.
TbD is composed of a set of visual reasoning primitives relying on attention transformations, allowing the model to perform reasoning by composing attention masks.
The authors compare the accuracy of their model tested on CLEVR, alongside with several existing models. In particular, they show improvement over the previously mentioned PG + EE, CNN + GRU + FiLM and MAC models.

They also compare their performance on CoGenT against PG+EE.
Nonetheless, the description lacks information about the used sets for training, fine-tuning and testing. 
They indicate that they used 3k images and 30k questions from the CoGenT Condition B for fine-tuning of their model, but do not explicitly point the set they used the samples from.

\subsubsection{Our MAC and S-MAC models and methodology}

Due to the fact that the test sets ground truth labels for both CLEVR and CoGenT aren't publicly available, we decided to follow the approach proposed in~\cite{perez2017film} and split the CoGenT B Validation set.
As a result we used 30k samples for fine-tuning and the remaining ones for testing.
When testing the model on CoGenT A, we used the whole validation set.

Moreover, as we were using the validation sets for testing, we also splitted the training set and used 90\% for training and the remaining 10\% for validation during training.
We used the same methodology for all the results presented in this paper, i.e. \tableref{tab:results}, \tableref{tab:generalization_comparison} and \tableref{tab:results_full}.

%% file: conclusion.tex
\section{Conclusion}
\label{sec:conclusion}

We have introduced S-MAC, a simplified variant of the MAC model. Because it has nearly half the number of parameters in the recurrent portion MAC cell, it trains faster while maintaining an equivalent test accuracy. 

Our experiments on zero-short learning show that the MAC model has poor performance in line with the other models in the literature. Thus, this remains an interesting problem to investigate how we can train it to disentangle the concepts of shape and color.

With fine-tuning, the MAC model indeed achieves much improved performance, matching state of the art. However, we have showed that correlation between the different domains must be taken into account when fine-tuning, otherwise potentially leading to decreased performance.

%% file: appendix.tex
\appendix

 \section{Full MAC and S-MAC comparison}
\label{sec:full_comparison}

In \tableref{tab:results_full} we present the full comparison between MAC and S-MAC models achieved with our implementations of both models.
In the [Row] column we indicate the measures that we have  analyzed and discussed in the experiments section of the main paper.

\begin{table}[!h]
	\centering
	\begin{tabular}{ccccCcCcc}
		\toprule
		\multirow{2}{*}{Model} & \multicolumn{3}{c}{Training} &  \multicolumn{2}{c}{Fine-tuning} & \multicolumn{2}{c}{Test} & \multirow{2}{*}{Row} \\
		\cmidrule{2-4} \cmidrule{5-6} \cmidrule{7-8}
		& Dataset                & Time [h:m] & Acc [\%]          & Dataset & Acc [\%]  & Dataset & Acc [\%] & \\
		\midrule
		\multirow{15}{*}{MAC} & \multirow{10}{*}{CLEVR}  & \multirow{10}{*}{30:52}  & \multirow{10}{*}{96.70} & \multirow{4}{*}{--}   & \multirow{4}{*}{--}  & CLEVR    & 96.17    & (a)      \\
		\cmidrule{7-8} 
		&                        &  &               &     &                                & CoGenT-A    &  96.22 &  \\
		\cmidrule{7-8} 
		&                        &   &              &     &                               & CoGenT-B   & 96.27 & \\
		
		\cmidrule{5-6} \cmidrule{7-8} 
		&                             &                                         &    &   \multirow{2}{*}{CoGenT-A}         &       \multirow{2}{*}{98.06}          & CoGenT-A &  94.60	   &      \\
		\cmidrule{7-8} 
		&                             &                                         &       &         &                & CoGenT-B &    93.28   &    \\
		\cmidrule{5-6} \cmidrule{7-8} 
		&                             &                                         &    &   \multirow{2}{*}{CoGenT-B}         &       \multirow{2}{*}{98.16}          & CoGenT-A &  93.02    &     \\
		\cmidrule{7-8} 
		&                             &                                         &       &         &                & CoGenT-B &    94.44   &    \\  
		
		\cmidrule{2-4} \cmidrule{5-6} \cmidrule{7-8} 
		& \multirow{5}{*}{CoGenT-A} & \multirow{5}{*}{30:52}     & \multirow{5}{*}{97.02}   &  \multirow{2}{*}{--}  &  \multirow{2}{*}{--}    & CoGenT-A & 96.88    &     \\
		\cmidrule{7-8} 
		&                             &                                         &       &         &                & CoGenT-B & 79.54     &    \\
		\cmidrule{5-6} \cmidrule{7-8} 
		&                             &                                         &    &   \multirow{2}{*}{CoGenT-B}         &       \multirow{2}{*}{97.91}          & CoGenT-A &  92.06     &    \\
		\cmidrule{7-8} 
		&                             &                                         &       &         &                & CoGenT-B &    95.62   &    \\
		\midrule
		\multirow{15}{*}{S-MAC} & \multirow{10}{*}{CLEVR}  & \multirow{10}{*}{28:30}  & \multirow{10}{*}{95.82} & \multirow{3}{*}{--}   & \multirow{3}{*}{--}  & CLEVR    & 95.29     & (b)      \\
		\cmidrule{7-8} 
		&                        &  &               &     &                                & CoGenT-A    &  95.47 & (d)  \\
		\cmidrule{7-8} 
		&                        &   &              &     &                               & CoGenT-B   &  95.58 & (e) \\		
		
		\cmidrule{5-6} \cmidrule{7-8} 
		&                             &                                         &    &   \multirow{2}{*}{CoGenT-A}         &       \multirow{2}{*}{97.48}          & CoGenT-A &  93.44      &   \\
		\cmidrule{7-8} 
		&                             &                                         &       &         &                & CoGenT-B &    92.31   &    \\
		\cmidrule{5-6} \cmidrule{7-8} 
		&                             &                                         &    &   \multirow{2}{*}{CoGenT-B}         &       \multirow{2}{*}{97.67}          & CoGenT-A &  92.11     & (i)    \\
		\cmidrule{7-8} 
		&                             &                                         &       &         &                & CoGenT-B &    92.95  & (j)     \\  		
		
		\cmidrule{2-4} \cmidrule{5-6} \cmidrule{7-8} 
		& \multirow{5}{*}{CoGenT-A}   & \multirow{5}{*}{28:33}   & \multirow{5}{*}{96.09}  &  \multirow{2}{*}{--}  &  \multirow{2}{*}{--}   & CoGenT-A & 95.91    & (c)      \\
		\cmidrule{7-8} 
		&                             &                                         &     &          &                & CogenT-B & 78.71       & (f)   \\
		\cmidrule{5-6} \cmidrule{7-8} 
		&                             &                                         &    &   \multirow{2}{*}{CoGenT-B}         &       \multirow{2}{*}{96.85}          & CoGenT-A &  91.24   & (g)      \\
		\cmidrule{7-8} 
		&                             &                                         &       &         &                & CoGenT-B &    94.55   & (h)    \\
		\bottomrule
	\end{tabular}
	\caption{CLEVR \& CoGenT accuracies for the MAC \& S-MAC models.}
	\label{tab:results_full}
\end{table}